\documentclass[default,iicol]{sn-jnl}

\usepackage{subdef}
\usepackage{graphicx}%
\usepackage{multirow}%
\usepackage{array}
\usepackage{multicol}
\usepackage{amsmath,amssymb,amsfonts}
\usepackage{caption}%
\usepackage{amsthm}%
\usepackage{mathrsfs}%
\usepackage[title]{appendix}%
\usepackage{xcolor}%
\usepackage{textcomp}%
\usepackage{manyfoot}%
\usepackage{booktabs}%
\usepackage{algorithm}%
\usepackage{algorithmicx}%
\usepackage{algpseudocode}%
\usepackage{listings}%
\usepackage[section]{placeins}
\usepackage{subfigure}
\usepackage{soul}
\usepackage{orcidlink}

\theoremstyle{thmstyleone}%
\theoremstyle{thmstyletwo}%

\theoremstyle{thmstylethree}%

\begin{document}


\title[\htr{ }]{\textbf{\framework{}: A Page-Level Handwritten Text Recognition System for Indic Scripts}}


\author[1]{Badri Vishal Kasuba\orcidlink{0000-0003-2636-7639}}
\equalcont{These authors contributed equally to this work.}
\author[1]{Dhruv Kudale\orcidlink{0000-0002-3862-819X}}
\equalcont{These authors contributed equally to this work.}

\author[1]{Venkatapathy Subramanian\orcidlink{0000-0002-4851-628X}}
\author[1]{Parag Chaudhuri\orcidlink{0000-0002-1706-5703}}
\author[1]{Ganesh Ramakrishnan\orcidlink{0000-0003-4533-2490}}
\affil[1]{\orgdiv{Department of Computer Science and Engineering}, \orgname{IIT Bombay}, \country{India}}


\email{\{badrivishalk, dhruvk, venkatapathy, paragc, ganesh\}@cse.iitb.ac.in}




\abstract{
In recent years, the field of Handwritten Text Recognition (\htr) has seen the emergence of various new models, each claiming to perform competitively better than the other in specific scenarios. However, making a fair comparison of these models is challenging due to inconsistent choices and diversity in test sets. Furthermore, recent advancements in \htr{} often fail to account for the diverse languages, especially Indic languages, likely due to the scarcity of relevant labeled datasets. Moreover, much of the previous work has focused primarily on character-level or word-level recognition, overlooking the crucial stage of Handwritten Text Detection (\htd) necessary for building a page-level end-to-end handwritten OCR pipeline. Through our paper, we address these gaps by making three pivotal contributions. Firstly, we present an end-to-end framework for \textbf{P}age-\textbf{L}evel h\textbf{A}ndwri\textbf{TT}en T\textbf{E}xt \textbf{R}ecognition (\framework{}) by treating it as a two-stage problem involving word-level \htd{} followed by \htr. This approach enables us to identify, assess, and address challenges in each stage independently. Secondly, we demonstrate the usage of \framework{} to measure the performance of our language-agnostic \htd{} model and present a consistent comparison of six trained \htr{} models on ten diverse Indic languages thereby encouraging consistent comparisons. Finally, we also release a Corpus of Handwritten Indic Scripts (\dataset), a meticulously curated, page-level Indic handwritten OCR dataset labeled for both detection and recognition purposes. Additionally, we release our code and trained models, to encourage further contributions in this direction.
}

\maketitle

\section{Introduction}\label{sec1}
\begin{figure*}
\includegraphics[width=\textwidth]{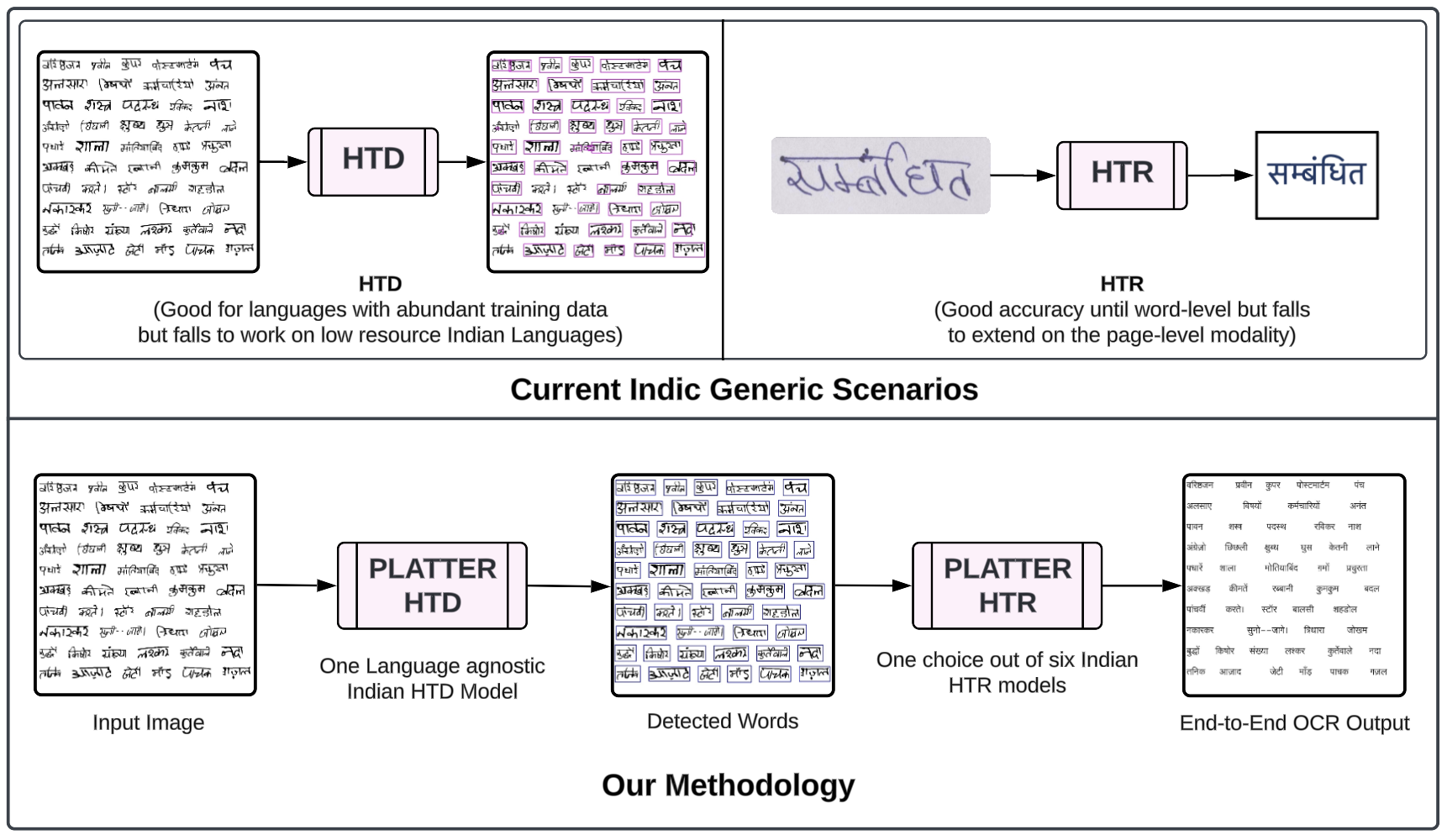}
\centering
\caption{Limitations of current HTD and HTR methods used for OCR of Indic languages are shown in the top row. We depict the schematic pipeline of \framework{} for end-to-end page-level OCR below.}
\label{fig:teaserfigure}
\end{figure*}

Handwritten OCR from document pages involves two main stages namely \htd{} and \htr{}. \htd{} is one of the very first stages in document analysis and the OCR workflow. It can be used to identify word-level handwritten textual content from a page. The output of \htd{} can later be used for different downstream tasks. On the other hand, \htr{} involves determining the corresponding textual information from images detected using \htd{} in a machine-readable format. \htr{} is beneficial for various domains including but not limited to document analysis, information retrieval, visual question-answering, etc. Handwritten documents are a rich source of information having many styles and strokes that come in various scripts. There are several handwritten documents having content written in Indic languages, however, digitizing these handwritten documents is more challenging due to the complex nature of Indic scripts. Hence, for effective digitization and editing, Indic Handwritten OCR proves to be a relevant problem even today. Besides that, the \htr{} problem has seldom been explored in Southeast Asian languages, unlike the abundant work done in English, Chinese, and other Latin-based languages. India is home to many handwritten texts, palm-leaf documents, imprinted texts on rocks, manuscripts, etc. Hence, developing such systems that can accurately detect and recognize handwritten text in Indic languages through an end-to-end pipeline would be useful in various tasks like cross-lingual information retrieval, translation to and from Indic languages, and so on.  

Fig \ref{fig:teaserfigure} highlights recent generic approaches used for \htd{} and \htr{} models. This is a depiction of a scenario about Indic languages. As shown in Fig \ref{fig:teaserfigure}, due to scarcity of labeled data, certain \htd{} models are unable to perform well on Indic language text detection. Indic languages are written in several different scripts, including Devanagari, Bengali, Tamil, Telugu, Kannada, Malayalam, and Odia, among others. Each script has its own set of alphabets. Some alphabets can be shared across scripts with modifications to suit the language. Though many Indic scripts share a common origin, these scripts evolved independently, leading to distinct variations in shapes and forms of alphabets. Further, multiple vowel and consonant combining rules lead to even more variations in these alphabets due to the presence of compound words and agglutination. The Unicode Standard has incorporated the alphabets of various Indic scripts under the group of Asian scripts. A good Indic \htd{} system must be robust enough to handle the high frequency and intensity of agglutination in Indic languages. However, Indic languages have huge vocabularies and a lot of conjunct characters, and when expressed in a handwritten modality have huge forms of variations as well depending on the writers' stokes and styles. Efforts must be made to develop Deep Learning (DL) based approaches to perform consistent word-level language-agnostic detection of handwritten text. Further, as seen in Fig \ref{fig:teaserfigure}, a good Indic \htr{} system can interpret the correct sequence of character representations from the word-level handwritten input image. However, it is necessary to extend their use to more practical page-level inputs. This could be achieved in two ways. One of them is working in conjunction with a \htd{} model which will detect and feed the word-level detections from a page to \htr{} model. The other is training a \htr{} model for page-level inputs itself. However, the latter poses a great amount of challenges which include dealing with the vocabulary of multiple languages, implicit script detection, the inability to decouple stages, and so on. There has been a tremendous boom in DL-based methods, especially consisting of Transformer-based models, to bring about effective word-level \htr{}. The use of vision-based transformers together with powerful and enhanced decoding mechanisms have assisted in solving word-level \htr{} problems to a considerable extent. These new word-level \htr{} architectures can prove to be beneficial in the OCR of handwritten text in documents that are written by various people, in several scripts, styles, and formats. Thus, we choose to look at this problem as a sequence of two independent stages of language-agnostic \htd{} followed by powerful word-level \htr{}. When these stages are used together sequentially, they can help us to design powerful handwritten OCR systems. As shown in Fig \ref{fig:teaserfigure}, our methodology in \framework{} of decoupling this end-to-end OCR pipeline as a two-stage problem helps not only to identify challenges in individual stages but also to address them independently. Developing robust \htd{} models capable of handling diverse scripts, alongside Indic \htr{} systems that accurately predict character sequences, is an essential part of designing \framework{}. Hence, through this paper, we try to reduce the gaps encountered in designing new systems that assist in performing Indic Handwritten OCR. Our proposition is \framework{}, a framework that will perform end-to-end Indic handwritten OCR by enabling users to make a custom choice of language-specific \htr{} models. We train and establish the use of a single relevant language-agnostic \htd{} model for detecting word-level images from an input page composed of handwritten text. We also adapt a few word-level recognition models for \htr{} tasks. We demonstrate the use of \framework{} to perform end-to-end Indic Handwritten OCR. We also provide qualitative and quantitative analyses of the performances of various possible combinations of \htr{} models. Our main contributions can be summarized as:

\begin{itemize}
    \item Proposition of \framework{}, an end-to-end Indic Handwritten OCR framework that enables users to select \htr{} model of choice. 
    \item We present the qualitative and quantitative analysis of various \htd{} and \htr{} models used within our \framework{} framework.
    \item Creation of \dataset{}, a large Indic Handwritten OCR page-level dataset that can be used to pre-train different models as well as avail benchmark scores.
    \item Releasing the relevant code-base here\footnote{https://github.com/iitb-research-code/doctr} and all the finetuned \htd{} and \htr{} models that were used in the analysis.
\end{itemize}

 The following Section \ref{sec:relwork} throws light on recent related work carried out in the fields of \htd{} and \htr{} respectively. Section \ref{sec:expt} explains how we have experimented with different datasets and how we have equipped our selected candidate architectures for \htd{} and \htr{} tasks. In Section \ref{sec:framework}, we provide a brief description in terms of the design and working of our framework, along with demonstrating the different use cases that \framework{} provides for further analysis. Eventually, we use our framework to report their performance and comparative analysis in Section \ref{sec:results}. We finally conclude our work and open doors for new contributions in Section \ref{sec:concl}.

\section{Related Literature}
\label{sec:relwork}

This section highlights different approaches introduced in recent literature that have been used for \htd{} and \htr, particularly for Indic scripts.

\subsection{Literature for \htd}
\label{sec:relwork-htd}
Initially, \htd{} was tackled either by simple Computer Vision (CV) algorithms or classical ML-based methods. These methods include edge detection \cite{sobel}, connected component analysis \cite{koo2013scene}, and contour-based methods \cite{contour}. All these methods have been used successfully for text detection tasks which are independent of the underlying scripts of the text. However, these CV-based methods have also been sensitive to noise. These limitations in terms of accuracy and robustness have been addressed by several DL-based methods. Differential Binarization Network (DBNet) \cite{dbnet} was introduced for generic scene text detection. DBNet is a segmentation-based detection network with a RESNET backbone, making it faster, more accurate, and lightweight to detect text in the form of bounding boxes. Since it is end-to-end trainable, it can be equipped well for \htd{} tasks. LinkNet \cite{linknet} attempts to exploit parameters utilization of neural networks efficiently. Similarly, other region-based and end-to-end trainable object detection networks can be used for \htd{} tasks. These methods use region proposal networks (RPNs) to identify regions of interest. Examples of such methods include object detection networks like Faster R-CNN \cite{rcnn} and Mask R-CNN \cite{maskrcnn}. However, all these DL-based methods heavily rely on labeled data. Since there is also a scarcity of labeled datasets consisting of Indic handwritten text, inference of object detection-based text detection networks show poor performance in detecting Indic handwritten words. On the other hand, some kinds of approaches pose the problem of text detection as that of minimization of a regression-based loss, through which coordinates of text regions are directly determined as predictions. They do not use region proposals and are typically faster than region-based methods. Regression-based text detection methods include TextBoxes \cite{textboxes}, EAST \cite{east}, and TextBoxes++ \cite{textboxes++}. However, regression-based methods have proven to be incapable of detecting curved text detection owing to their limitation to work on strictly rectangular bounding boxes. Another class of text detection methods includes pixel-wise classifications done through segmentation. The broad methodology for this includes performing segmentation of the input image into text and non-text regions. The output segmentation mask is then utilized to extract the text regions. PixelLink \cite{pixellink}, SegLink \cite{seglink}, and TextSnake\cite{textsnake} are examples of such segmentation-based methods. The main advantage of segmentation-based methods is that they can even detect arbitrarily shaped text very well, but they require a complex post-processing step that takes a considerable amount of inference time. 

Apart from that, various new DL-based architectures have been introduced like FCENET\cite{FCENet}, DBNet++\cite{dbnet++}, PSENet \cite{PSENet}, etc. that have shown to be effective on generic scene text detection that can be adapted to \htd{} tasks in the presence of relevant datasets. 
SeamFormer \cite{ic13-seamformer} attempts to perform line-level text detection on page images composed of manuscripts. 
Recently, with the success of various Transformer-based Object detection approaches such as DETR\cite{detr}, they have also been used in many language-agnostic text detection tasks. Currently, the DPText-Detr \cite{dptext} can capture the global relationships between image features and text features to carry out text detection from images. There has been an emergence of a lot of text detection datasets released for printed text and scene text detection. These datasets include English ICDAR15 \cite{icdar15}, Multilingual ICDAR17 \cite{icdar17}, SynthText \cite{synthtext}, etc. Very few handwritten texts like IAM \cite{iam} datasets are available for detection either in isolation or from entire page-level images. This is because most of the handwritten text datasets have been annotated for character-level or word-level \cite{bengali-char-data} keeping end-to-end tasks (with recognition) in mind. PHD-Indic-11 \cite{phd-indic} is a page-level script detection dataset for Indic languages. Further, there are very few labeled page-level Indic \htd{} datasets like the Bangla Handwriting dataset \mbox{\cite{majid}} which has character-level bounding boxes annotated for essay pages by several writers. TEXTRON \cite{textron} address the scarcity of Indic \htd{} datasets using a data programming-based paradigm. Thus, through this presented overview of various text detection methods, we adapt and justify the usage of DBNet for \htd{} as an integral part of our end-to-end Indic Handwritten OCR framework (\framework).

\subsection{Literature for \htr}
Initial work for offline \htr{} included rule-based approaches like pattern matching \cite{pal-old} which were dominated mostly by CV-based approaches. Such approaches mostly dealt with character-level inputs. Later ML-based work primarily focused on feature extraction was leveraged for \htr{}. Examples of such work include Kannada \cite{kannada} and Gujarati \cite{gujarati} handwritten character recognition. However, most of these ML-based approaches focused either on one language at a time or on character-level recognition. Character level recognition does not capture the presence of conjunct characters as observed in several Indic languages. Hence, the further space was dominated by word-level recognition. This area of \htr{} was captured by purpose-specific DL models like CRNN-VGG \cite{crnn-vgg} or incorporating Encoder-Decoder-based systems \cite{hocr-dl, iiit-improved}. CRNN-VGGs \cite{crnn-vgg}, CRNN with MobileNet backbones \cite{mobilenets} are end-to-end trainable networks for image-based sequence recognition. The handwritten OCR \cite{hocr-dl} has a CNN backbone to understand the visual and textual features. Show, Attend, and Read model (SAR) \cite{sar} uses off-the-shelf neural network components CNNs and LSTMs similar to CRNN but with an additional attention mechanism to create an end-to-end trainable model that does not require pre-training.

However, using transformers has paved the way for boosting \htr{} performance. Leveraging the concept of attention for image-to-sequence tasks, proper fine-tuning, and initialization \cite{ic13-htr-trans} has enabled Transformer-based architectures to provide state-of-the-art performances in \htr{}. Recently, a unified architecture composed of transformer-based encoder \cite{ic13-urdu-htr} has been employed to perform \htr{} in the Urdu language. Some page-level HTR approaches include Faster DAN \cite{ic13-dan-english-pl} for English and a real-world case study \cite{ic13-hebrew-pl} performed for page-level images composed of resource-scarce language texts. MASTER \cite{master} tries to leverage the concept of global (non-local) attention to address the drawbacks of traditional image-based sequence recognition approaches like CRNN-VGG \cite{crnn-vgg}. To incorporate global context-based attention, MASTER draws inspiration from GCNet \cite{gcnet}. PARSEQ \cite{parseq} leverages the use of Permutation Language Models (PLM) so that end-to-end training can be done on all possible sequences of predicted tokens. ViTSTR \cite{vit} and TROCR\cite{trocr} form a part of another end-to-end Transformer-based OCR model that uses a pre-trained VIT \cite{vit} model to extract visual features from images at the encoding stage and then uses a relevant decoding mechanism to decode the corresponding text that needs to be recognized. Apart from recent work, various Indic \htr{} dataset releases and competitions \cite{htr-competition} have been taking place which include PBOK \cite{pbok}, ROYDB \cite{roydb}, TAMIL-DB \cite{tamildb} etc. One such notable work includes the proposition of a benchmark dataset of \textbf{IIIT-Indic-HW-Words} \cite{iiit-indic} along with CRNN-based \cite{crnn-vgg} baseline models for each language to perform \htr{}. The dataset consists of handwritten word-level images of over 1 million images covering 10 Indic languages namely Bengali, Gujarati, Gurumukhi, Hindi (Devanagari), Kannada, Malayalam, Odia, Tamil, Telugu, and Urdu. As mentioned previously, the Bangla Handwriting dataset \mbox{\cite{majid}} includes characters annotated from word and character essays. More such datasets include character-level Devanagari \cite{devanagari_handwritten_character}, word-level Bangla \cite{bangla_word_level}, IIIT-HW-Dev \cite{iiit-dev}, IIIT-HW-Tel \cite{iiit-tel}, etc. Besides the datasets, there has also been a recent emergence of various tools required to annotate Indic handwritten text. The Automatic Annotation Tool \cite{annot-tool} also demonstrates the process of annotating handwritten data for 13 Indic languages. Further, there have also been proposed architectures like ScrabbleGAN \cite{scrabblegan} that can be used to generate synthetic handwritten text images. Several recent \htr{} dataset generation approaches like \cite{ic13-htr-data-style} try to induce writing style in the images of handwritten text to make it appear more realistic. More recently released approaches like AFFGAN \cite{ic13-htr-data-affgan} and DDPM-based generated training samples \cite{ic13-htr-data-ddpm} generate synthetic data for \htr{} tasks. One use of synthetic data is seen in dataset augmentation for Tamil \htr{} \cite{tamil} suggesting that training data when augmented with synthetic images can boost the performance of \htr{} models.

\section{Experimentation}
\label{sec:expt}
This section presents an overview of datasets used within our experiments to equip different architectures for different \htd{} and \htr{} tasks.

\subsection{Datasets}
Considering the case of Handwritten OCR in Indic languages, there has always been a scarcity of labeled and benchmark data. For curating page-level annotated and benchmark data, we have used benchmark IIIT-Indic-HW-Words dataset\cite{iiit-indic} which is annotated for word-level \htr{} in ten Indic languages. To study and train \htd{} models we have also used a subset of the PhD-Indic11 dataset \cite{phd-indic}. We have also created synthetic page-level data through ScrabbleGAN \cite{scrabblegan} to augment training data for \htr{} models. We describe the dataset statistics and present how we have used them in the following subsections.

\subsubsection{IIIT-Indic-HW-Words}
\label{sec:dataset-iiit}
\begin{figure*}[htbp]
    \centering
    \begin{minipage}[t]{0.48\textwidth}
        \centering
        \includegraphics[width=\linewidth]{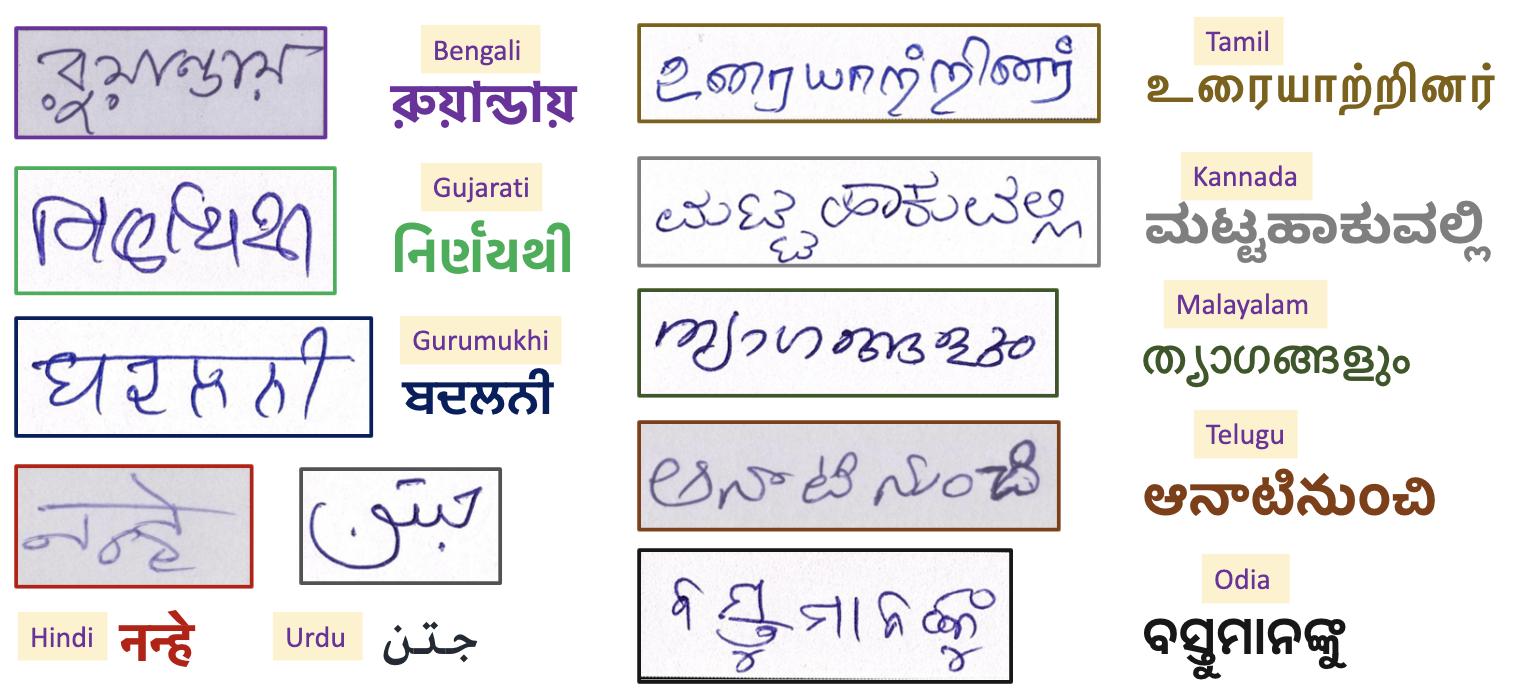}
\caption*{IIIT-Indic-HW-Words} 
    \end{minipage}\hfill
    \begin{minipage}[t]{0.5\textwidth}
        \centering
        \includegraphics[scale =0.17]{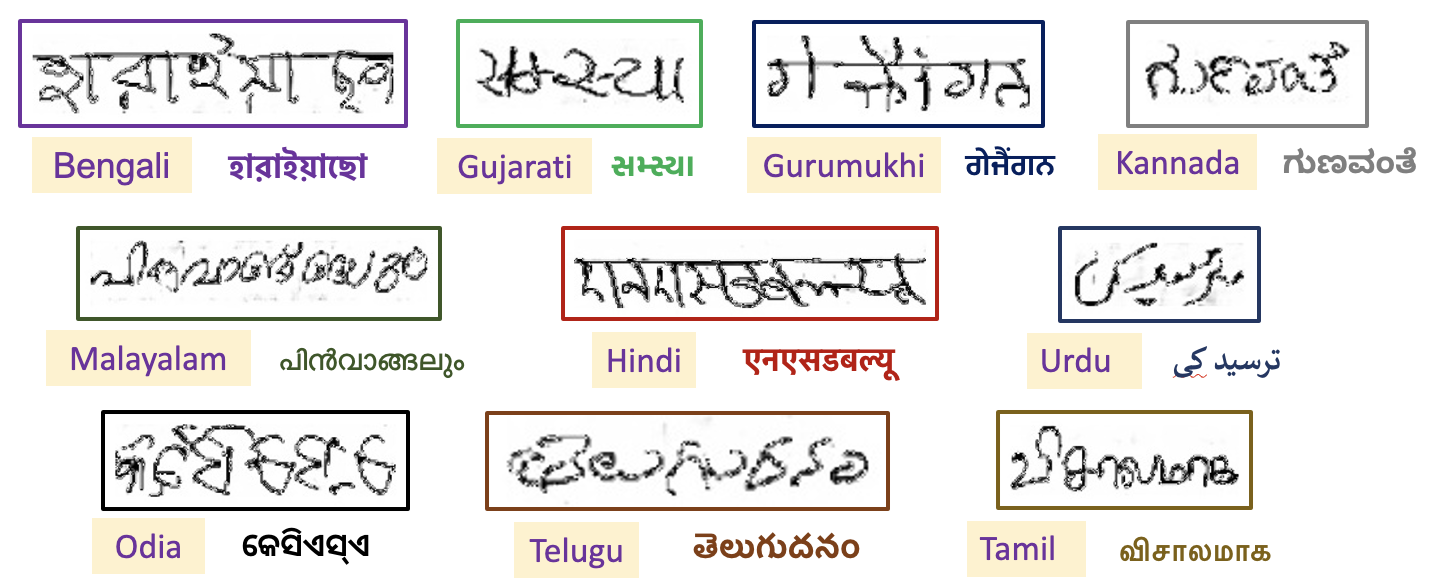}
\caption*{Synthetic word-level dataset} 
\label{fig:synth_images}
    \end{minipage}

\caption{Sample word-level images of handwritten text from different datasets in various Indic languages}
\label{fig:word_samples}
\end{figure*}

IIIT-Indic-HW-Words dataset is a benchmark word-level \htr{} dataset that is first-of-its-kind to open-source large data for Indic model development. The dataset contains $1,081,730$ word-level images covering $10$ Indic languages. The various languages and their corresponding train, val, and test sets are highlighted in Table \ref{tab:indic}. These images have been curated through the manual writing of several annotators in the $10$ languages. We use this dataset to provide performance analysis for various \htr{} models in an isolated format. We have also used this dataset to create our page-level dataset \dataset{} for further experimentation on the development of our end-to-end \framework{} framework. Fig \ref{fig:word_samples} highlights a few samples from this dataset.

\begin{table}[h]
\centering
\caption{IIIT-Indic-HW-Words statistics}
\begin{tabular}{|c|c|c|c|c|}
\hline
\multirow{2}{*}{{ \textbf{Language}}} &  \multicolumn{3}{c|}{{\textbf{Total Words}}}\\
 & Train  & Val & Test \\
\hline
Bengali    & 82,554 & 12,947 & 17,574   \\
Gujarati   & 82,563 & 17,643 & 16,490   \\
Gurumukhi  & 81,042 & 13,627 & 17,947   \\
Hindi      & 69,853 & 12,708 & 12,869   \\
Kannada    & 73,517 & 13,752 & 15,730   \\
Malayalam  & 85,270 & 11,878 & 19,635   \\
Odia       & 73,400 & 11,217 & 16,851   \\
Tamil      & 75,736 & 11,598 & 16,184   \\
Telugu     & 80,637 & 19,980 & 17,898   \\
Urdu       & 71,207 & 13,906 & 15,517   \\ \hline
Total      & 775,779 & 139,256 & 166,695   \\
\hline
\end{tabular}

\label{tab:indic}
\end{table}

\subsubsection{Synthetic Word-level Dataset}
\label{sec:dataset-synth}
We have used the ScrabbleGAN \cite{scrabblegan} model to synthesize word-level handwritten images for all $10$ Indic languages. ScrabbleGAN is a semi-supervised generative model designed to synthesize handwritten text images that exhibit versatility in both style and lexicon. Operating in a semi-supervised manner, ScrabbleGAN utilizes both labeled and unlabeled data, offering a significant improvement over fully supervised methods. One of the unique features of ScrabbleGAN is the ability of its generator to manipulate the resulting text style. We use this synthetic data for the augmentation of existing training sets. This approach allows any \htr{} model trained on augmented data to adapt to unseen images during inference. We have trained the ScrabbleGAN model for each of the ten Indic languages on respective IIIT-Indic-HW-Words for $250$ epochs. Using this, we synthesized word-level images. Around $500,000$ unique handwritten word-level images have been synthesized for 10 languages each. The generation of synthetic data helped in cutting the cost of training the model from scratch and in the faster convergence of the \htr{} model training as we demonstrate in Section \ref{sec:htr}. Table \ref{tab:synth} describes the details of this synthetically generated word-level dataset.

\begin{table}[h]
\caption{Synthetic word-level dataset statistics}
\begin{tabular}{|c|c|c|}
\hline
{ \textbf{Language}} & { \textbf{Words}} & { \textbf{Pages}}  \\
\hline
Bengali   & 500,000   & 11,399    \\ 
Gujarati  & 500,000   & 10,175    \\
Gurumukhi & 500,000   & 09,826     \\ 
Hindi     & 500,000   & 10,791    \\
Kannada   & 500,000   & 13,341    \\ 
Malayalam & 500,000   & 12,106    \\
Odia      & 500,000   & 10,247    \\ 
Tamil     & 500,000   & 10,901    \\
Telugu    & 500,000   & 11,879    \\ 
Urdu      & 500,000   & 10,562     \\ \hline
Total     & 5,000,000  & 111,227 \\     
\hline
\end{tabular}
\label{tab:synth}
\end{table}

\subsubsection{PHD-Indic-11}
\label{sec:dataset-phd}
The PHD-Indic11 dataset was originally developed for script identification over 11 languages. It is a page-level handwritten dataset composed of text written by different annotators. We have annotated word-level bounding boxes on each required page having Indic language content through various annotators and thus curated this text detection dataset. The word-level annotation statistics of handwritten pages in this are mentioned in Table \ref{tab:phdindic}. As a part of our work, we have used the annotated pages of $10$ Indic languages as a subset of a larger training set to train our fine-tuned \htd{} model(s). Fig \ref{fig:combined} shows sample images of a few pages from this dataset. We describe more about training \htd{} models in Section \ref{sec:htd}.

\begin{table}[h]
\caption{Number of pages and words in the PhDIndic11 dataset}
\begin{tabular}{|c|c|c|}
\hline
\textbf{Language} & \textbf{Pages} & \textbf{Word-boxes} \\ 
\hline
Bengali   & 161 & 13,627 \\
Gujarati  & 100 & 16,000 \\
Gurumukhi & 132 & 17,355 \\
Hindi     & 220 & 25,264 \\
Kannada   & 046 & 01,794  \\
Malayalam & 107 & 05,038  \\
Odia      & 172 & 12,482 \\
Tamil     & 120 & 05,504  \\
Telugu    & 085 & 05,503  \\
Urdu      & 201 & 15,381 \\ \hline
Total     & 1,344 & 117,948 \\
\hline
\end{tabular}
\label{tab:phdindic}
\end{table}

\subsubsection{\dataset{} dataset}
\label{sec:dataset-chips}
\begin{figure}
\centering
\includegraphics[width=0.45\textwidth]{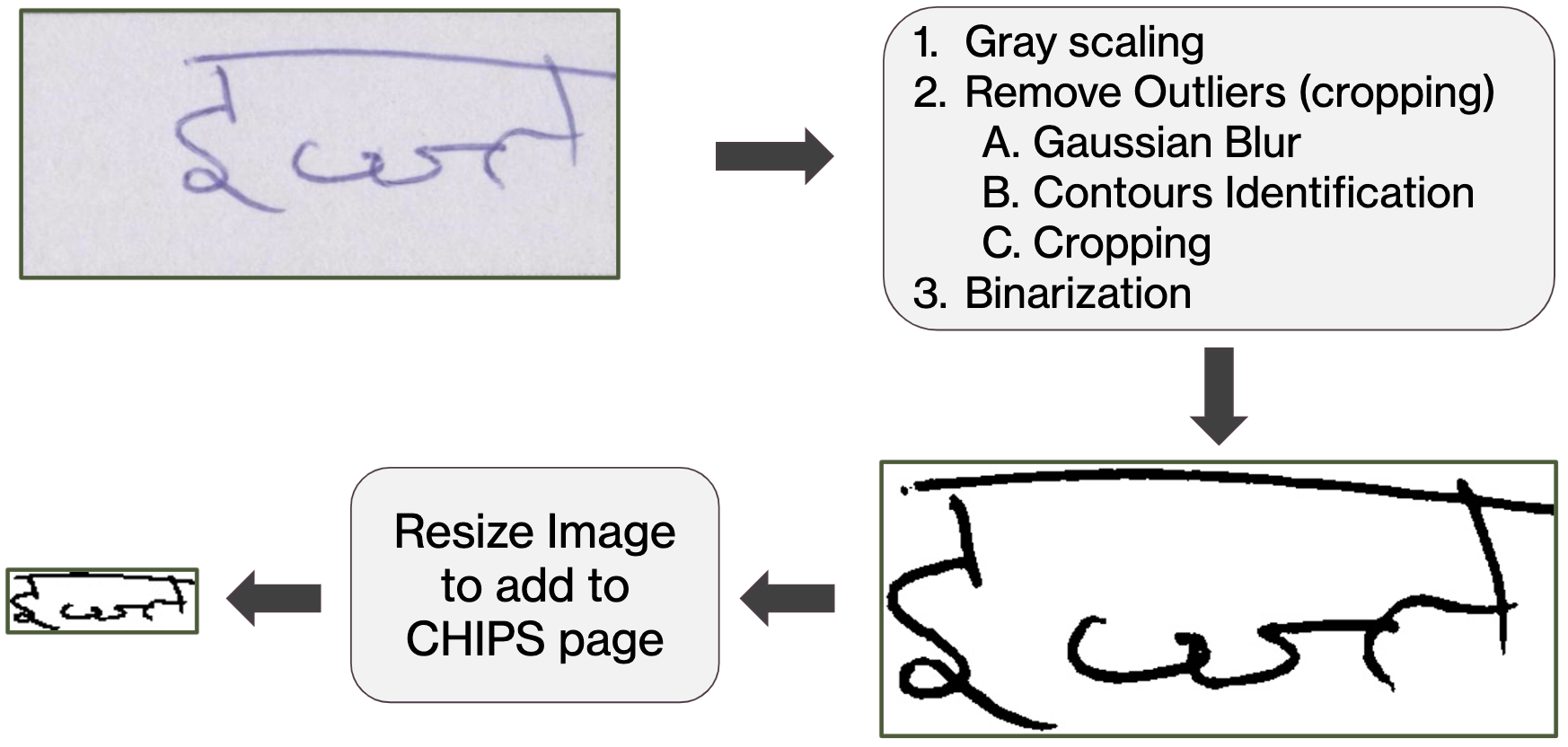}
\caption{Image pre-processing for generating word-level images to be incorporated in \dataset{}}
\label{fig:chips-preprocess}
\end{figure}

\begin{figure}
\centering
\includegraphics[width=0.3\textwidth]{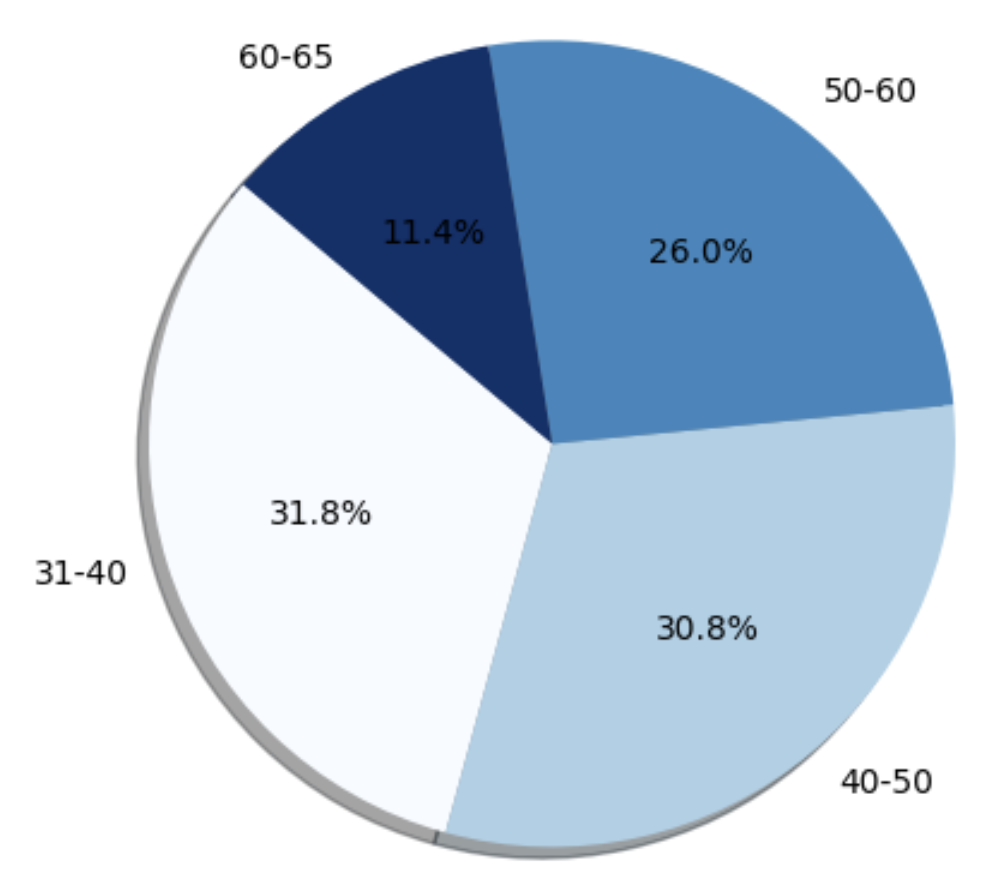}
\caption{Distribution of different word heights (in pixels) for words used in \mbox{\dataset{}} pages}
\label{fig:fonts}
\end{figure}

As a part of the generation of \dataset{}, a Corpus of Handwritten Indic scripts, we have utilized the IIIT-Indic-HW-Words dataset as mentioned in Section \ref{sec:dataset-iiit}. We have appended the labeled word-level images from the IIIT-Indic-HW-Words dataset to fill in a $1024*1024$ sized page image, each image having different word heights. The word height is defined by the height of the preprocessed image constituting the word to be added on the \dataset{} page. The preprocessing algorithm to convert a raw IIIT-Indic-HW-Words image into a suitable one to add on a \dataset{} page is highlighted in Fig \mbox{\ref{fig:chips-preprocess}}. As shown, first, we convert the input image into a grayscale image and then apply Gaussian blur to eliminate small noises. Further, we detect contours to isolate the text. Before downsizing, we apply dilation to the contours to preserve all textual information. Any small noises outside the main contour are also removed. Apart from the small noises, we also try to remove extra ruled lines from word-level images during the preprocessing stage. We start scanning the image from the extreme ends and check if the number of non-white pixels (pixels that constitute writing strokes) per column is less than a certain threshold which we have set as 10 pixels. Then, if the value is less than the threshold, we identify that column to be a part of the horizontal ruling line and hence crop that column out of the word box. We use a similar procedure row-wise to remove the extra vertical ruled lines. Thus, we finally crop out the region constituting the word and binarize it to be added to the final \dataset{} page. As highlighted in Fig \mbox{\ref{fig:chips-preprocess}}, the next stage carries out resizing of the cropped word image. As \htd{} and \htr{} are prone to changing word sizes, we have created \dataset{} pages having words with varying heights. Further, we also ensure that the word-level images in the same line of \dataset{} page have variable heights that vary only from 0.8 to 1.2 times the reference word height of the page. To mimic actual handwritten text on a page, every page has a fixed reference word height which was set randomly to any number between 32 and 64. The distribution of different word heights used in the \dataset{} pages is shown in Fig \mbox{\ref{fig:fonts}}. 

\begin{figure*}[htbp]
    \centering
    \begin{minipage}[t]{0.47\textwidth}
        \centering
        \includegraphics[width=\linewidth]{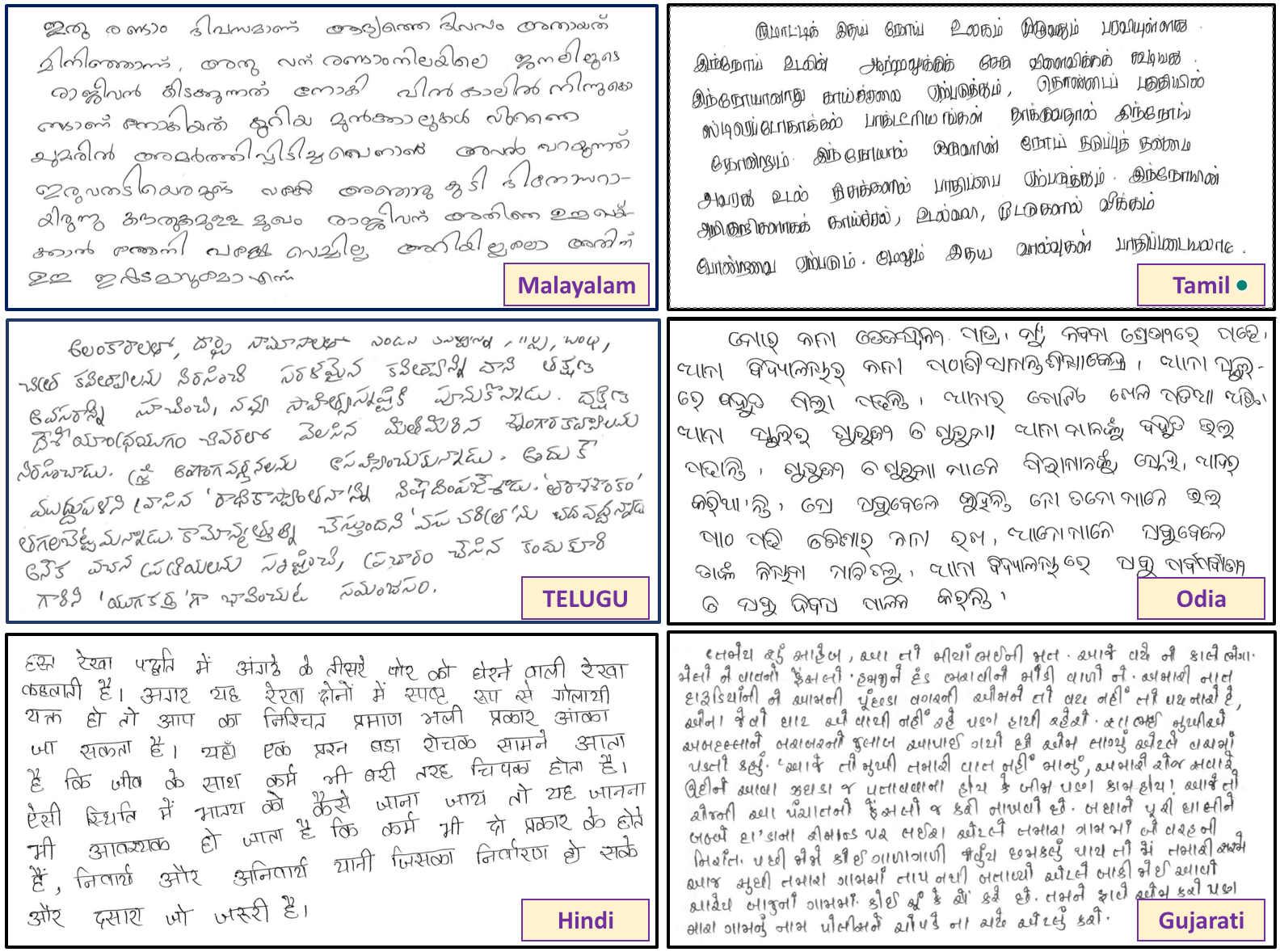}
        \caption*{PhD-Indic11 Dataset} 
        \label{fig:phd-indic}
    \end{minipage}\hfill
    \begin{minipage}[t]{0.5\textwidth}
        \centering
        \includegraphics[scale=0.16]{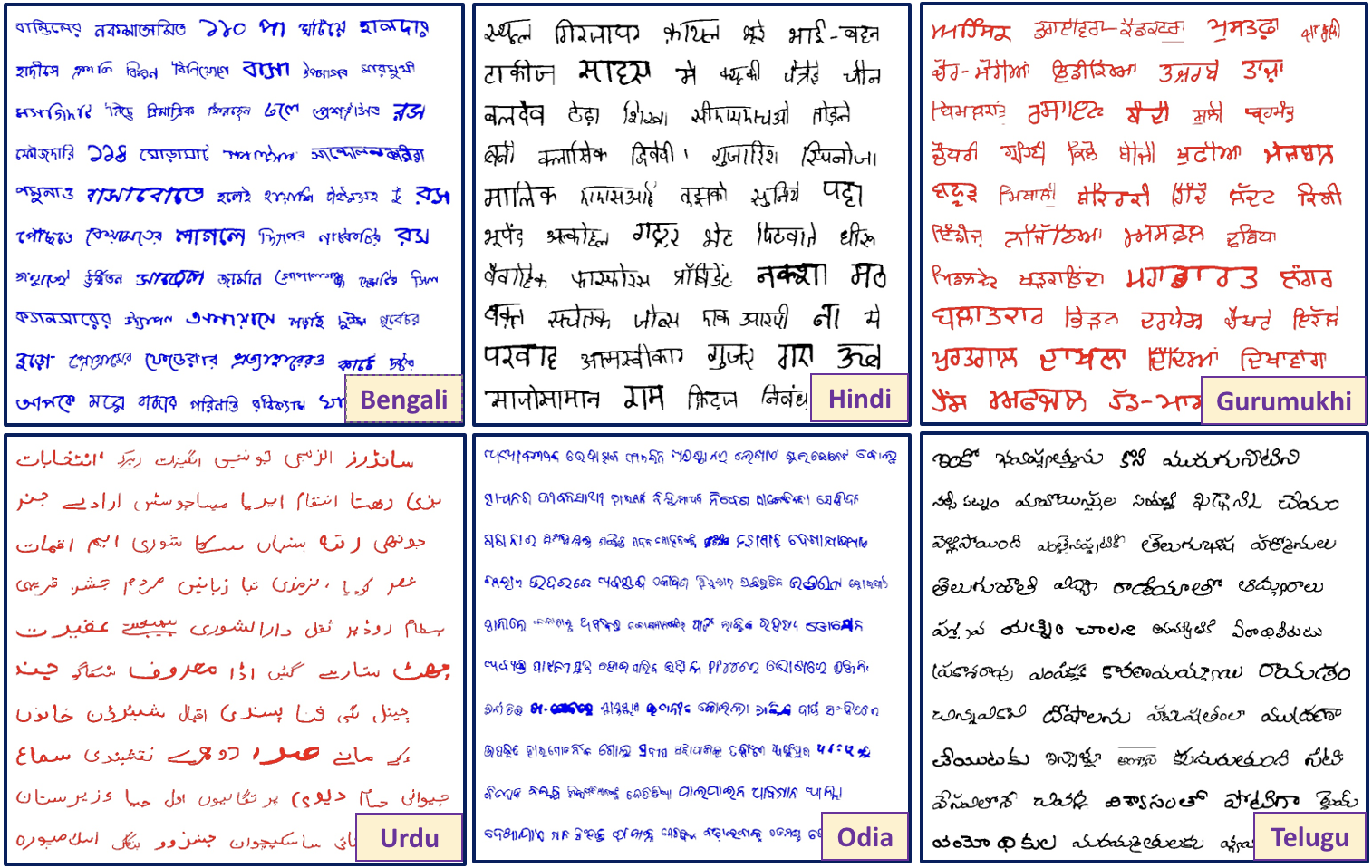}
        \caption*{\dataset{} dataset}
        \label{fig:chips-sample}
    \end{minipage}
\caption{Sample page-level images of handwritten text from different datasets in various Indic languages}
\label{fig:combined}
\end{figure*}

We also ensured that every word underwent preprocessing and resizing to fit perfectly within the $1024*1024$ page without overflowing along with appropriate spacing between them. This was governed by parameters enlisted in Table \ref{tab:params_pagelevel}. The parameters could be manually tuned based on requirements. \textit{PAGE-WIDTH} and \textit{PAGE-HEIGHT} parameters are passed to create the page image of the concerned size. We have created $1024*1024$ sized page-level data for our work. \textit{SPACE-X} and \textit{SPACE-Y} give the choice of altering the gap between words either randomly or to a fixed value. The default value for both is 32. However, for creating \dataset{} pages, we set \textit{SPACE-X} to randomly attain a value between 1 to 3 times the average character width. \textit{BORDER-CUT-X} and \textit{BORDER-CUT-Y} parameters are used to crop out the exterior border noise present in the word-level images before they enter the preprocessing stage in the horizontal and vertical directions respectively. This meticulous process ensured the generation of a high-quality, versatile dataset that could replicate handwritten data. Fig \ref{fig:combined} shows sample images of a few pages of the \dataset{} dataset. Table \ref{tab:page-level} contains the statistics of different languages in \dataset{}. \dataset{} could be used as a benchmark for page-level OCR. We also report our end-to-end scores for page-level handwritten OCR on \dataset{} test set. 

\begin{table}[]
\centering
\caption{Statistics of \dataset{}}
\begin{tabular}{|c|c|c|c|}
\hline
{ \textbf{Language}} & { \textbf{Train}} & { \textbf{Val}} & { \textbf{Test}} \\
\hline
Bengali   & 1,600  & 261  & 339  \\
Gujarati  & 1,603  & 347  & 323   \\
Gurumukhi & 1,354  & 226  & 290   \\
Hindi     & 1,123  & 187  & 209   \\
Kannada   & 1,969  & 372  & 424   \\
Malayalam & 3,419  & 470  & 784   \\
Odia      & 1,406  & 212  & 331   \\
Tamil     & 2,572  & 398  & 545   \\
Telugu    & 1,951  & 504  & 419   \\
Urdu      & 1,154  & 238  & 260   \\
\hline
Total     & 18,151 & 3,215 & 3,924   \\
\hline
\end{tabular}
\label{tab:page-level}
\end{table}

\begin{table}[h]
\renewcommand{\arraystretch}{1.2}
\caption{Default parameters used for \dataset{}}
\centering
\begin{tabular}{|c|c|}
\hline
\textbf{Parameter} & {\textbf{Values}} \\
\hline
\textit{REFERENCE-WORD-HEIGHT} & Range(32, 64)\\
\hline
\textit{PAGE-WIDTH * PAGE-HEIGHT} & $1024*1024$  \\ \hline
\textit{SPACE-X, SPACE-Y}   & 32, 32       \\ \hline
\textit{BORDER-CUT-X, BORDER-CUT-Y}  & 3.5, 3.5    \\ 
\hline
\end{tabular}
\label{tab:params_pagelevel}
\end{table}

\subsection{\htd{} Models}
\label{sec:htd}

\begin{table*}[!h]
\caption{Comparison of detection performance of ten individual language-specific models (columns 3, 4 and 5) against a single language agnostic model (columns 6, 7, and 8) on \dataset{} test sets for an IoU of 0.7}
\centering
\begin{tabular}{|c|c|ccc|ccc|}
\hline
{}  & {} & \multicolumn{3}{|c|}{{ \tiny{\textbf{Individual Language Fine-tuned}}}} & \multicolumn{3}{|c|}{{ \tiny{\textbf{Language Agnostic Fine-tuned}}}} \\
\multirow{-2}{*}{{ \textbf{Language}}} & \multirow{-2}{*}{{ \textbf{Pages}}} & P & R & F1 & P & R & F1 \\
\hline
Bengali   & 339  & 99.30 & 99.48 & 99.39 & 98.92 & 99.08 & 99.00  \\
Gujarati  & 323  & 99.32 & 99.45 & 99.38 & 99.25 & 99.35 & 99.30  \\
Gurumukhi & 290  & 98.76 & 98.88 & 98.82 & 97.49 & 97.90 & 97.69  \\
Hindi     & 209  & 98.20 & 98.29 & 98.25 & 97.54 & 97.69 & 97.62  \\
Kannada   & 424  & 99.43 & 99.59 & 98.25 & 98.68 & 99.30 & 98.99  \\
Malayalam & 784  & 98.76 & 99.15 & 98.96 & 97.37 & 98.76 & 98.06  \\
Odia      & 331  & 99.07 & 99.28 & 99.18 & 99.02 & 99.24 & 99.13  \\
Tamil     & 545  & 99.01 & 99.21 & 99.15 & 94.45 & 97.41 & 95.91  \\
Telugu    & 419  & 99.26 & 99.36 & 99.31 & 98.71 & 99.30 & 99.01  \\
Urdu      & 260  & 98.40 & 98.61 & 98.51 & 97.44 & 97.42 & 97.43  \\

\hline
Average   & 3,924 & \textbf{98.95} & \textbf{99.13} & \textbf{98.92} & 97.89 & 98.54 & 98.21   \\
\hline
\end{tabular}
\label{tab:detection}
\end{table*}

\begin{table*}[!h]
\caption{Overall results on \dataset{} test set of candidate \htd{} models. * indicates the mean of all ten language-specific models tested on individual language test sets. The last 2 rows present the scores of our finetuned language-agnostic model trained on 10\% and 100\% \dataset{} pages respectively.}
\resizebox{\textwidth}{!}{
\begin{tabular}{|c|ccc|ccc|ccc|}
\hline
{ } & \multicolumn{3}{|c|}{{ \textbf{IoU  = 0.5}}}  & \multicolumn{3}{|c|}{{ \textbf{IoU = 0.75}}} & \multicolumn{3}{|c|}{{ \textbf{IoU = 0.9}}} \\
\multirow{-2}{*}{{ \textbf{Model}}} & P & R  & F1 & P  & R & F1 & P  & R & F1   \\
\hline
Tesseract   & 66.58  & 83.09 & 73.93  & 4.01 & 5.0 & 4.45 & 0.01 & 0.02 & 0.01 \\  
Pre-trained DBNet  & 84.63  & 92.17 & 88.24   & 59.31 & 64.59 & 61.84 & 12.26 & 13.35 & 12.78 \\
*Language Specific DBNet   & 99.49 & 99.68 & 99.59   & 97.13 & 97.32 & 97.23 & 46.96 & 47.05 & 47.01 \\
Language agnostic DBNet - 10\% & 98.84 & 99.54 & 99.19 & 96.32 & 96.99 & 96.65 & 49.40 & 49.50 & 49.45  \\
Language agnostic DBNet - 100\% & \textbf{99.60} & \textbf{99.80} & \textbf{99.70} & \textbf{97.39} & \textbf{97.59} & \textbf{97.49}  & \textbf{49.60} & \textbf{49.95} & \textbf{49.77}  \\
\hline
\end{tabular}
}
\label{tab:detection-overall}
\end{table*}

\begin{figure*}[!h]
\centering
\subfigure{
\includegraphics[width=0.22\textwidth]{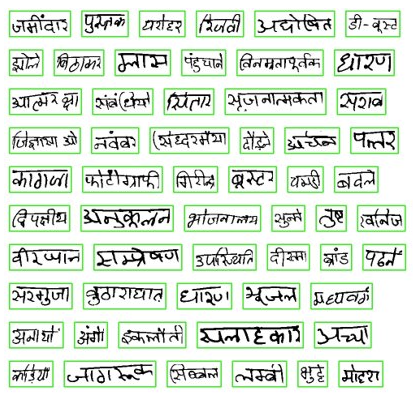} 
}
\subfigure{
\includegraphics[width=0.22\textwidth]{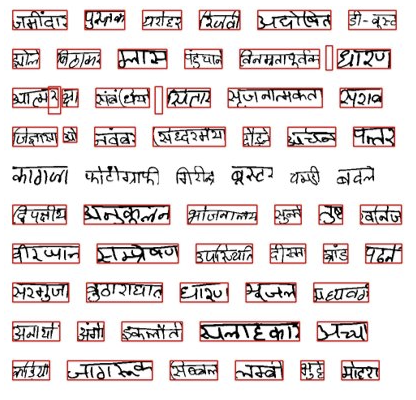} 
}
\subfigure{
\includegraphics[width=0.22\textwidth]{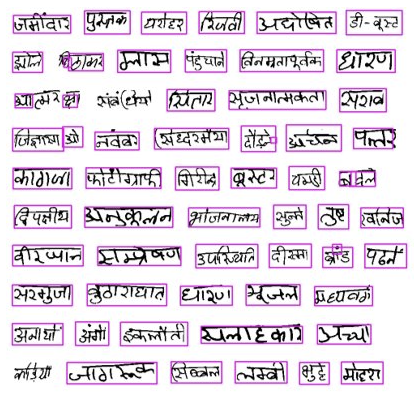}
}
\subfigure{
\includegraphics[width=0.22\textwidth]{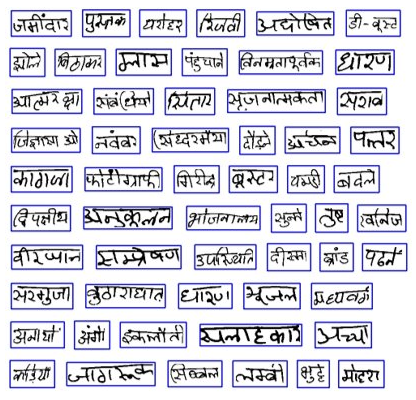}
}

\caption{Bounding boxes engulfing the words of a single Hindi page-level image produced by (L to R) ground truths, Tesseract, pre-trained DBNet, and our language agnostic fine-tuned DBNet respectively}
\label{fig:Detection preds}
\end{figure*}

\htd{} is the initial stage for page-level OCR. The bounding boxes of the detected words on the page are passed on to the \htr{} stage. Most of the state-of-the-art text detection models are DL-based models as mentioned in Section \ref{sec:relwork-htd}. Hence, training relevant \htd{} models, especially for identifying Indic texts requires a lot of labeled data. We use \dataset{} and the PHDIndic-11 dataset having rectangular bounding boxes as corresponding labels for page-level images for training a suitable \htd{} model. After initial survey and experimentation, we believe that DBNet \cite{dbnet} can be suitable for detecting words for our case. DBNet internally works on a pixel-based classification that can handle Indic script word detections. It is end-to-end trainable as well as a lightweight model which will benefit \framework{} framework. To ensure a resilient model for \htd{}, the question that arises is whether we should train individual models for distinct languages (\textit{n} models for \textit{n} languages), or should we pursue a single language-agnostic model. To answer this, we have evaluated both scenarios. We trained 10 language-specific DBNet models for 10 languages using the \dataset{} training sets. We also train a single language-agnostic DBNet model for all ten languages on a 10\% mixed subset of \dataset{} training images. The mixed subset incorporated all ten languages and was specifically chosen to compare accurately with language-specific models. The performance of both is compared in Table \ref{tab:detection}. We report Precision, Recall, and F1 Score for the performances of ten language-specific DBNet models and the single language-agnostic DBNet model on the \dataset{} test set for an IoU threshold of 0.7. We also mention the average of both the approaches along with a comparison to pre-trained DBNet and Tesseract in Table \ref{tab:detection-overall}. Our presented results show a very negligible difference in the performance of the language-specific model(s) and language-agnostic model. Since \framework{} is a collection of multiple models, we favor relying on a single language-agnostic model to reduce latency and overhead for the end-to-end OCR pipeline. Thus, we strongly advocate for our language-agnostic \htd{} model to be used in any generic OCR framework. Fig \ref{fig:Detection preds} shows the page-level predictions of a single page obtained using different \htd{} models.

\subsection{\htr{} Models}
\label{sec:htr}

\htr{} stands as a pivotal stage in the continuum of Handwritten OCR. Unlike detection, which can be language agnostic and operate on page-level input, HTR is closer to linguistic properties. An ideal \htr{} model needs to properly interpret a word image to generate appropriate character sequences. However, when we talk about the complexity of Indic languages, they are characterized by extensive vocabularies and very diverse character sets. This requires specialized language-specific (or script-specific) training of \htr{} models (unlike the language agnostic models for \htd). In addressing this need, our study dives into six distinct \htr{} models. They comprise three CRNN-based models namely CRNN-VGG \cite{crnn-vgg}, MobileNet \cite{mobilenets}, and SAR \cite{sar} along with three transformer-based models which include MASTER \cite{master}, ViTSTR \cite{vit}, and PARSEQ \cite{parseq}. These models are strategically chosen for their potential adaptability to the task of word-level Indic \htr{} as pointed out in the recent survey \cite{htr-survey}. For our analysis, we train these six \htr{} models using the DocTR \footnote{DocTR framework \url{https://github.com/mindee/doctr}} framework. During training the CRNN-based architectures, we have kept the default configs unchanged. We have trained the CRNN-VGG for 500 epochs and MobileNet and SAR were trained for 200 epochs. The learning rate for all three approaches was kept at 0.001. Transformer-based architectures require more epochs to converge. All ten language-specific models each for PARSEQ, ViTSTR, and MASTER were trained for around 500, 600, and 100 epochs respectively. While the constant learning rate for PARSEQ and ViTSTR was kept at 0.0001, it was kept at 0.001 for MASTER. To cover larger training samples and converge faster, we performed augmentation with synthetic word-level data. The different combinations of training sets and the number of epochs required to converge as well as the corresponding CRR and WRR on the Bengali test set are highlighted in Table \ref{tab:synth-analysis}. 

\begin{table}[]
\caption{Augmenting synthetic data to training pipeline for better generalization for the Bengali language. The 'Epochs' column stands for the number of epochs required for convergence of the CRNN model. The CRR and WRR represent the validation scores.}
\centering
\begin{tabular}{|c|c|c|c|}
\hline
{ \textbf{CRNN\_VGG Model}} & {\textbf{Epochs}} & {\textbf{CRR}} & {\textbf{WRR}} \\
\hline
SYNTH only  & 15  & 65.45  & 15.34    \\
Synth + 25\% train data  & 18  & 92.29  & 57.74\\
Synth + 50\% train data  & 39  & 94.63  & 68.08\\
Synth + 75\% train data  & 28  & 95.11  & 70.39\\
Synth + full train data  & 37  & 95.73  &73.47\\ 
\hline
\end{tabular}
\label{tab:synth-analysis}
\end{table}

\begin{table*}[]
\caption{The benefit of augmentation of Hindi training data expressed in terms of (i) the number of epochs required for training CRNN model to converge and (ii) the validation accuracy (CRR)}.
\centering
\resizebox{\textwidth}{!}{%
\begin{tabular}{|c|cc|cc|}
\hline
{} & \multicolumn{2}{|c|}{{ \textbf{IIIT-INDIC-HW Data Only}}} & \multicolumn{2}{|c|}{{ \textbf{Augmented IIIT-INDIC-HW Data}}} \\
\multirow{-2}{*}{{ \textbf{MODEL}}} & Epochs & Val Accuracy (\%) & Epochs & Val Accuracy (\%)\\
\hline
CRNN\_VGG  & 47  & 74.04 & 15  & 77.73 \\
MOBILENET  & 70  & 32.71 & 30  & 35.34 \\ 
SAR        & 37  & 74.39 & 12  & 81.81 \\
MASTER     & 20  & 85.25 & 10  & 82.67 \\
PARSEQ     & 33  & 84.21 & 17  & 73.35 \\
\hline
\end{tabular}
}
\label{tab:synth-benefit}
\end{table*}

\begin{figure}
\centering
\includegraphics[width=0.4\textwidth]{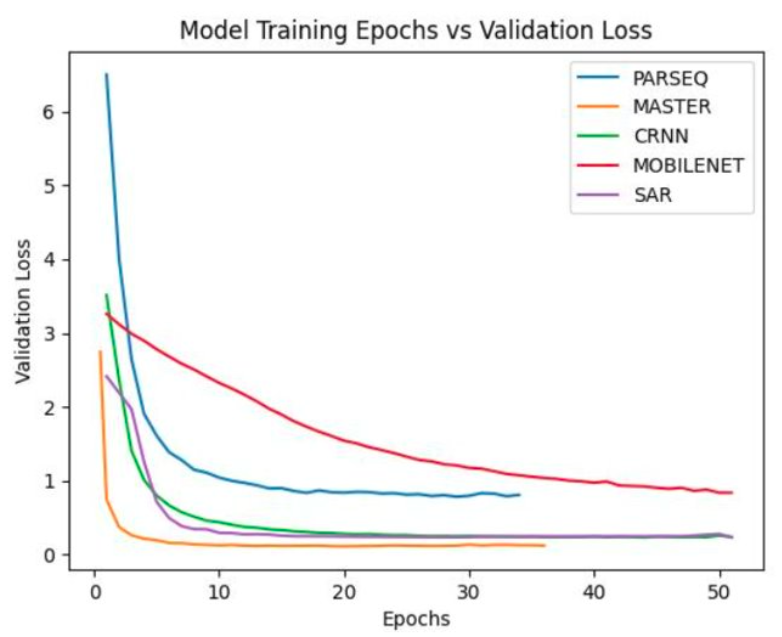}
\caption{Validation loss vs epochs for the training of various \htr{} models for the Bengali language} 
\label{fig:train_graph}
\end{figure}

Synthetic data augmentation to training data helped in faster training convergence giving rise to robust models as well. Hence, we performed a similar training experiment for the Hindi language with the same parameters. The augmented training dataset as described above was created by augmentation of Hindi words from IIIT-Indic-HW-Words \cite{iiit-indic} with the synthetic word-level data (described in Section \ref{sec:dataset-synth}) images of the same language. As seen in Table \ref{tab:synth-benefit}, we can see the faster convergence of models trained with augmentation of IIIT-Indic-HW-Words with synthetic data converged sooner (in fewer epochs) to yield slightly better validation accuracies. Hence, we adopted this kind of augmentation to train each of the six models for all ten Indic languages. All the learning rates finalized for training the final 60 models (6 models * 10 languages) were determined after extensive experimentation of training performance, validation accuracy, and convergence time. The training graph for all six models is depicted in Fig \ref{fig:train_graph}. All these models were trained on NVIDIA RTX A6000 GPU. Thus, using this comprehensive approach, we incorporated these trained models in \framework{}. We will now talk about the usage of these incorporated \htd{} and \htr{} models into our framework in the upcoming Section \ref{sec:framework}.

\section{\framework{} Framework}
This section describes the structure and workflow of \framework{}. We further also show how can \framework{} be used to carry out a wide variety of tasks including but not limited to page inference, output-based reconstruction, performance evaluation of different \htr{} models, latency analysis, and visualization of consistent comparisons of multiple models. 

\label{sec:framework}
\subsection{Design}
\begin{figure*}
\centering
\includegraphics[width=\textwidth]{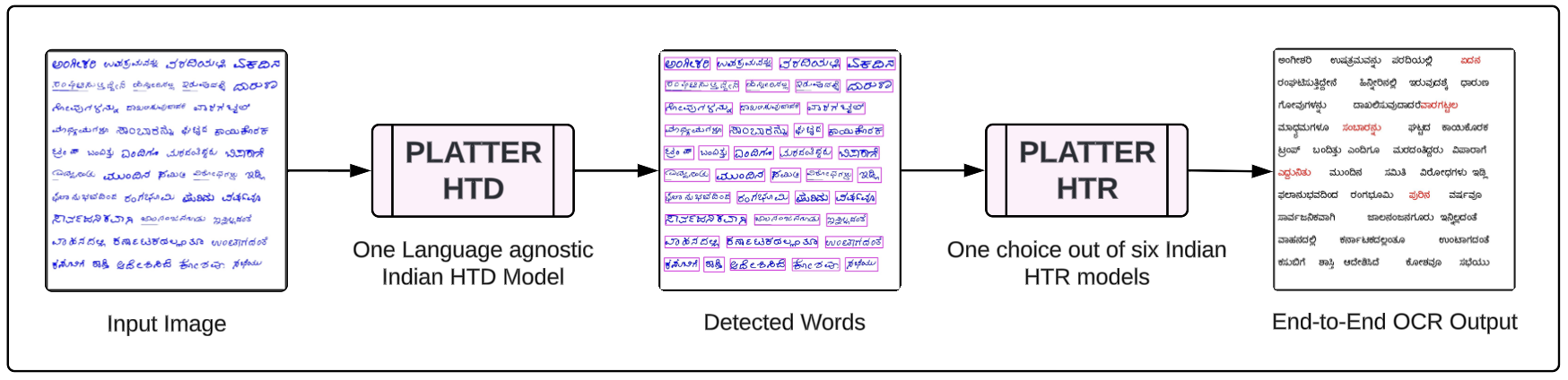}
\caption{Design of \mbox{\framework{}} illustrating the outputs of \htd{} and \htr{} stages respectively.}
\label{fig:platterdesign}
\end{figure*}

The flowchart explaining the structure and working of \framework{} is shown in Fig \ref{fig:platterdesign}. As seen, once the input page is passed onto the framework, our fine-tuned language agnostic \htd{} model detects the word-level bounding boxes present in the page image. Each word image is cropped and fed to the HTR model of choice. The language is a part of user input, however, the user gets a choice of one or more models out of the six \htr{} models to perform corresponding text recognition. Using \framework{}, we can thus carry out OCR of handwritten text for a page. We further demonstrate the usage of our tool in visualizing the qualitative and quantitative results respectively.

\begin{figure*}[]
\includegraphics[width=\textwidth]{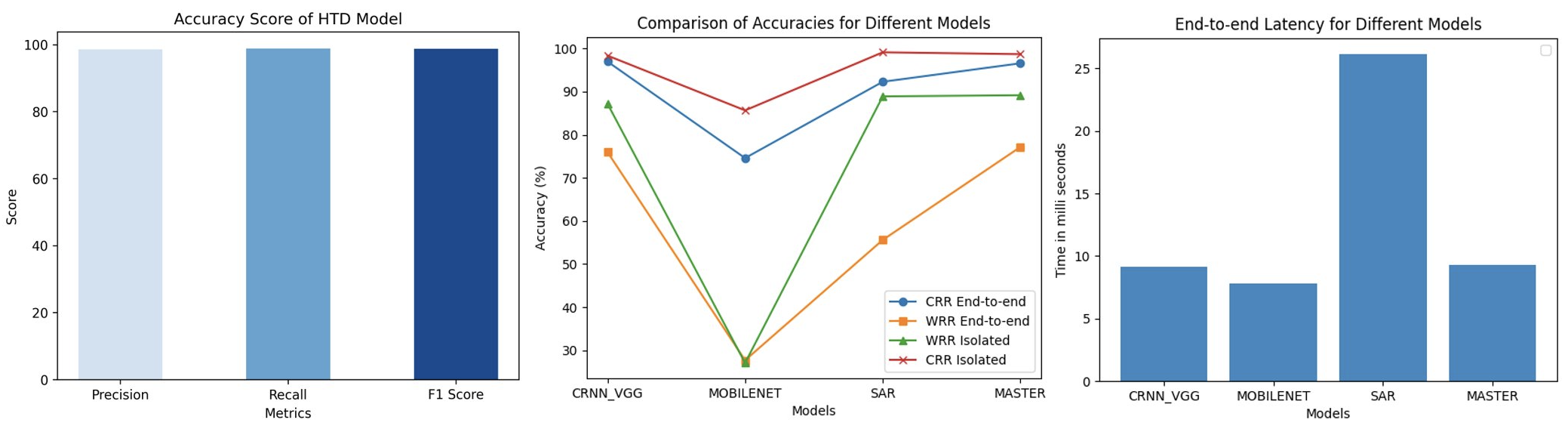}
\centering
\caption{\mbox{\framework{}} Graphs for visualization of comparison of language agnostic \htd{} and several \htr{} models chosen. The graphs (from L to R) represent our \htd{} performance, comparison of selected \htr{} models, and latency graphs for end-to-end OCR pipeline per word for the Malayalam \dataset{} test set.}
\label{fig:plattergraphs}
\end{figure*}

\subsection{Analysis}
Through this section, we demonstrate the usage of \framework{} for the following use cases about page-level handwritten OCR. 

\subsubsection{Single Page Visualization}
This is one of the most basic uses that performs OCR of a page image composed of multiple handwritten words. The framework takes a single image, the \htr{} model, and language as input. \framework{} sequentially performs \htd{} word-level inference and each cropped word image is provided to the \htr{} model reflecting the user's choice. Later the detected bounding boxes and recognized words are used to create an image containing the recognized words at proper positions. This helps us visualize the qualitative outputs of the end-to-end handwritten OCR pipeline. Fig \ref{fig:platterdesign} showcases the workflow of a single-page visualization. \framework{} can also provide outputs in text files or predefined formats like HOCR, JSON, etc.


\subsubsection{Performance Comparison}
\framework{} enables to provide the quantitative analysis of the entire end-to-end OCR pipeline on a custom-labeled set of images. This is how multiple \htr{} models can be consistently compared on the same test set.  This feature takes a folder path of a test set of page images (and corresponding labels) of the concerned language, and a set of one or more \htr{} models. Once analysis is performed, \framework{} displays the following details for the entire test set:

\begin{enumerate}
    \item \textbf{\htd{} Overall Performance} which mentions the Precision, Recall, F1 Score with a default IoU threshold of 0.5 used for detecting words in the page by our fine-tuned language agnostic \htd{} model.
    
    \item \textbf{Isolated CRRs and WRRs} for selected \htr{} models are displayed. This assists in making an overall comparison of the character recognition rate (CRR) as well as word recognition rate (WRR) of chosen \htr{} models for the particular language. The isolated metrics indicate that only the \htr{} model is assessed independently. The recognized words are compared against the ground truth words. Each word entry is defined by the bounding boxes that contain it, and are part of the true labels of the \dataset{} dataset.
    
    \item \textbf{End-to-end CRRs and WRRs} for selected \htr{} models are also mentioned. As we try to evaluate the end-to-end pipeline performance, these metrics take into account the result of using the language-agnostic \htd{} model on the input page image followed by \htr{} models of choice. In this scenario, the \htr{} model receives the word-level detections of our fine-tuned language agnostic \htd{} model. The recognized words are compared against the ground truth words, where the true word entry is identified by the bounding boxes of ground truths that share the maximum overlap with \htd{} detections.
    
    \item \textbf{Mean end-to-end latency}, which measures the average time taken per word for performing complete end-to-end OCR. One must note that, since the \htd{} model is common in all cases, the difference in latency is observed due to the choice of \htr{} models.
    
\end{enumerate}


\framework{} gives the option to visualize these results using which we get three graphs - one corresponding to the measures of \htd{} performance, another one comparing isolated and end-to-end accuracies of different \htr{} methods, and a third graph showing the mean end-to-end latency for all the chosen \htr{} models per word of the test set pages. The results are seen in Fig \ref{fig:plattergraphs}. Thus, using \framework{} for different scenarios on various test sets from \dataset{} we report the results and analyze performances in Section~\ref{sec:results}.

\section{Results and Discussions}
\label{sec:results}
This section shows the independent evaluation and analyses of different \htd{} and \htr{} models that have been made using \framework{}.

\subsection{\htd{} Performance}
In this section, we present the performance of our single fine-tuned language agnostic \htd{} model on all 10 Indic languages considered within our experimentation. We have trained this language agnostic \htd{} model on the complete (100\%) \dataset{} training set. All our further results have been reported using this \htd{} model. We report our results on the \dataset{} test set. Both the detections given by our \htd{} model and the ground truths are available in bounding boxes format. Table \ref{tab:detection-overall} from Section \ref{sec:htd} highlights the respective performances of our fine-tuned language agnostic \htd{} model for detecting handwritten text in different Indic languages.

\subsection{Isolated \htr{} Performance}
\begin{figure*}[h]
\includegraphics[width=0.8\textwidth]{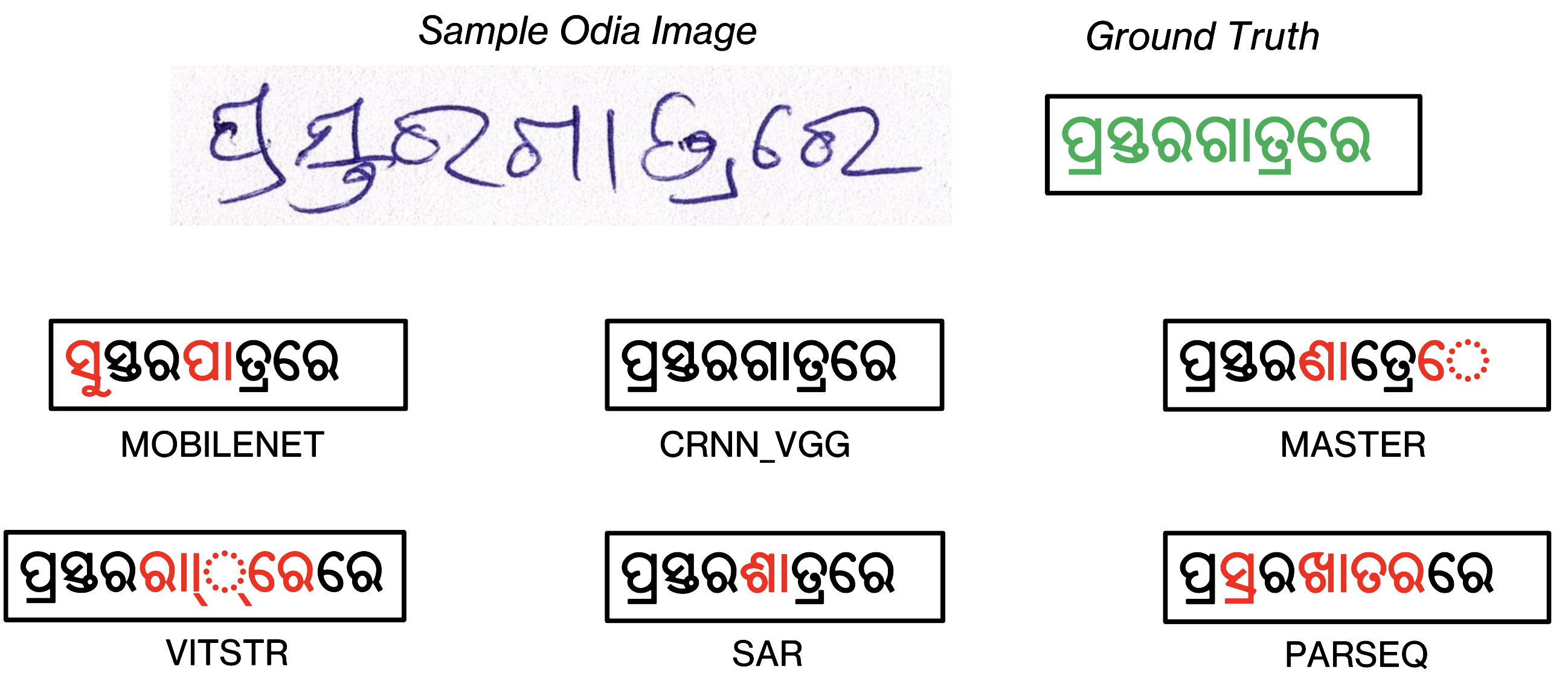}
\centering
\caption{Qualitative Recognition results representation for a sample isolated word-level input image on Odia language using different \htr{} models as represented in second and third rows The second and third rows, wherein the characters marked in red indicate an incorrect prediction.}
\label{fig:qual-analysis-isolated}
\end{figure*}

\begin{figure}[h]
\includegraphics[width=0.5\textwidth]{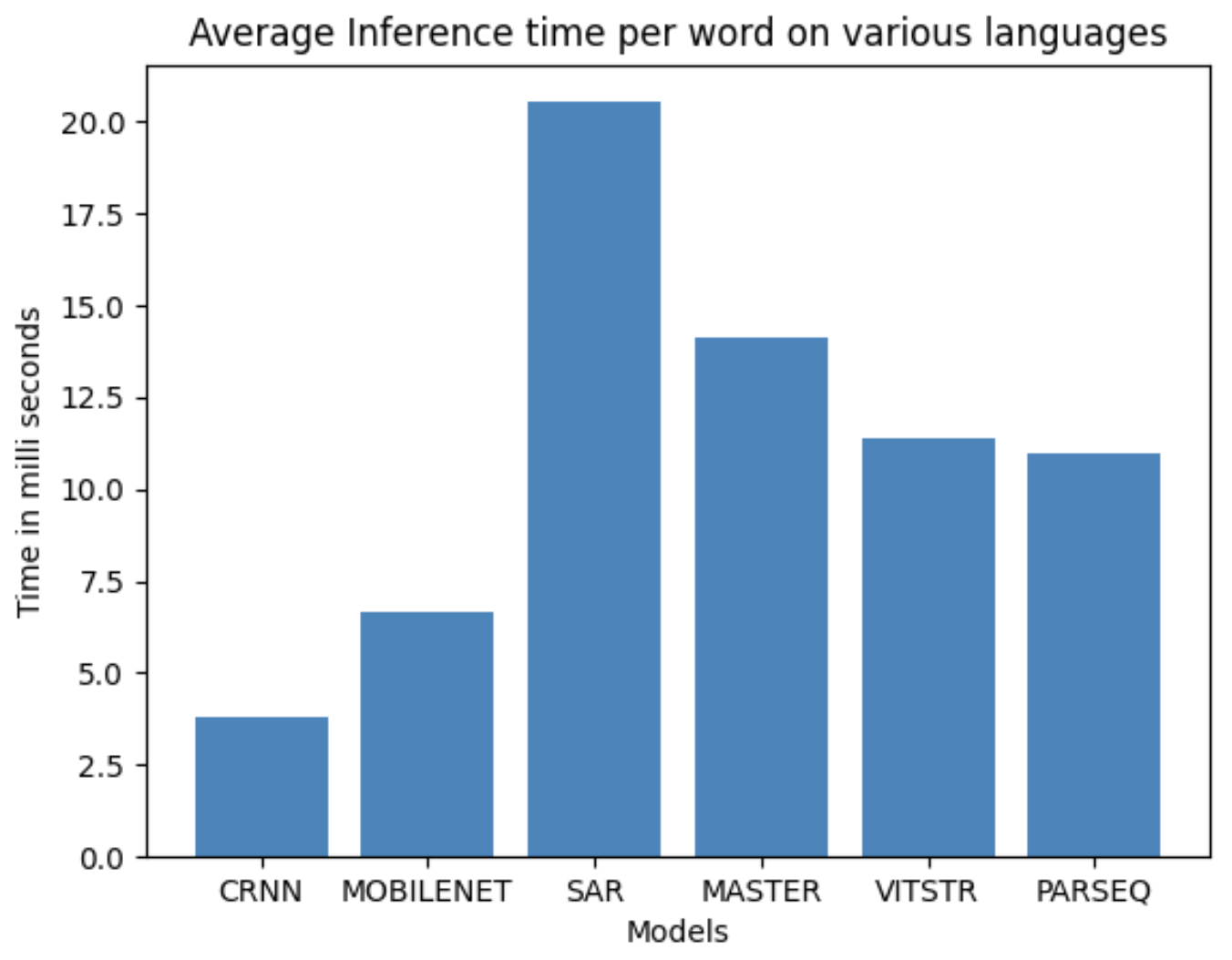}
\caption{Mean latency graph for the \htr{} models}
\label{fig:latencygraph}
\end{figure}

Table \ref{tab:isolated} highlights the performance of various \htr{} models on the \dataset{} test set on all ten languages. We present the CRR and WRR for evaluating isolated recognition performance for every language. As shown in the table, we can see that CRNN is the best-performing \htr{} model, when considering the isolated CRR, whereas PARSEQ takes over as the best-performing model in terms of isolated WRR. Further, Fig \ref{fig:qual-analysis-isolated} displays the output of all six \htr{} models on a single word-level image input. This gives an estimate of their qualitative outputs for a certain language. Besides that, we also provide an estimate of the latency of all \htr{} models per word for all ten languages in Fig \ref{fig:latencygraph}. Since for end-to-end OCR, the \htd{} stage uses a single model, we provide a comparative analysis of the processing time of different \htr{} models. A similar trend is observed in recording the latency for end-to-end OCR. CRNN is the fastest model among the presented six models whereas SAR is the slowest model for word-level inference.

\subsection{End-to-end Performance}
\begin{figure*}[h]
\centering
\subfigure{
\includegraphics[width=\textwidth]{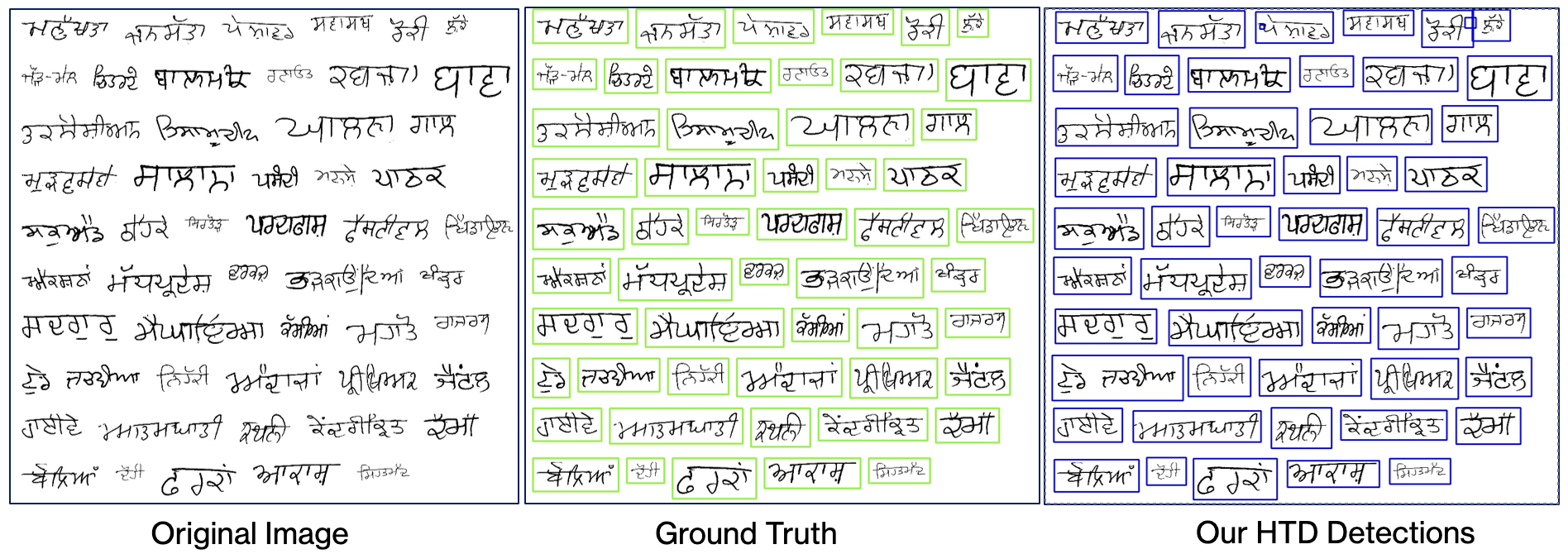} 
}
\subfigure{
\includegraphics[width=\textwidth]{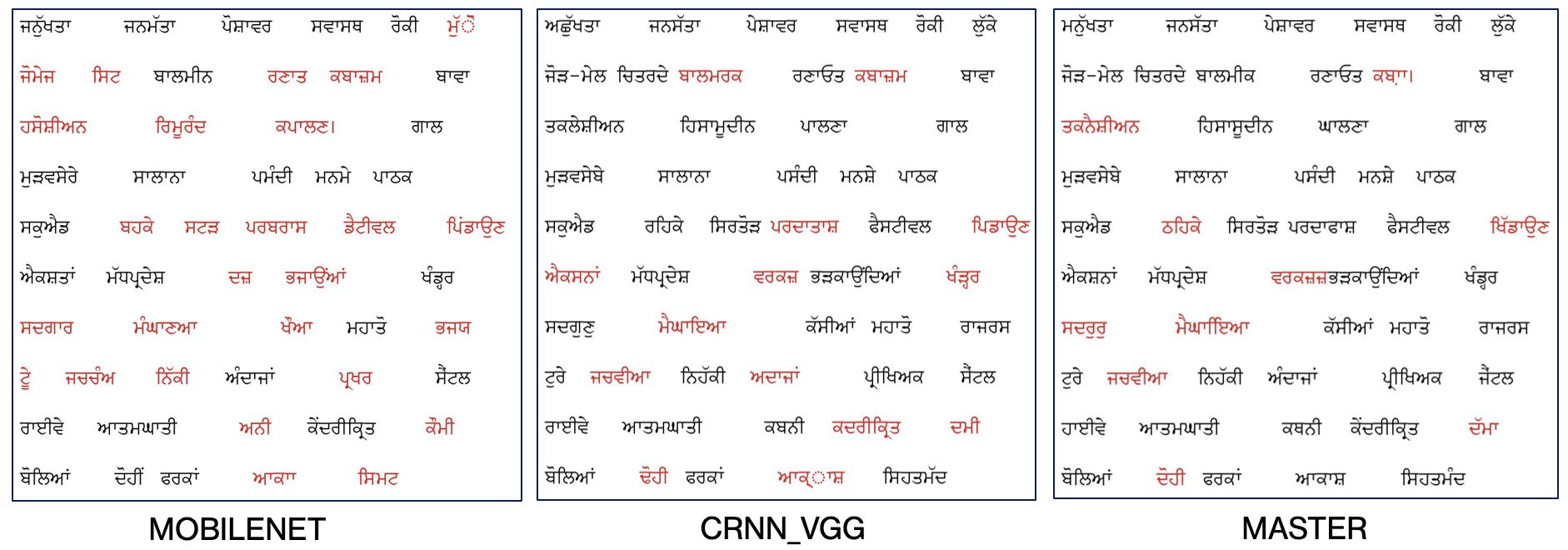} 
}
\subfigure{
\includegraphics[width=\textwidth]{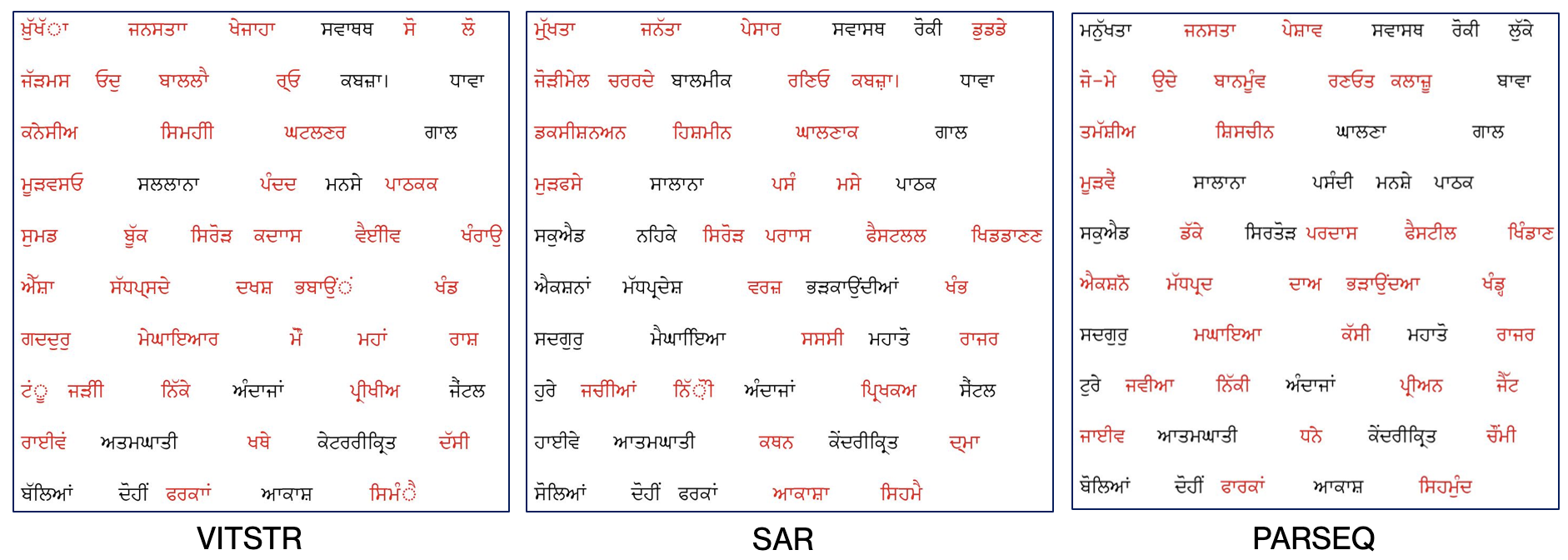} 
}
\caption{Outputs for end-to-end page level OCR using \framework{} for a single page having Hindi language text using different \htr{} models. The second and third rows represent the recognized outputs through different \htr{} models, wherein the words marked in red indicate incorrect predictions.}
\label{fig:qual-analysis-etoe}
\end{figure*}

In this subsection, we present the end-to-end performance on page-level data inputs. Table \ref{tab:end-to-end} highlights the performance of various \htr{} models on the \dataset{} test set for all 10 languages. To evaluate the end-to-end recognition performance, we compute CRR and WRR scores for every language from the predictions obtained by inferring the trained \htr{} models whose inputs are the word-level detections from our fine-tuned language agnostic \htd{} model. Fig \ref{fig:qual-analysis-etoe} displays the output of all six \htr{} models on a single page-level input image. As displayed in Table \ref{tab:end-to-end}, MOBILENET, SAR, and VITSTR models were not able to give comprehensive recognition results for the page-level dataset. Ironically, the PARSEQ model, which performs the best on isolated WRR, fails to perform well in end-to-end recognition. The PARSEQ model was trained on unprocessed IIIT-Indic-HW-Words images similar to other models but cannot generalize to cropped word-level inputs from \dataset{} pages. The CRNN and MASTER models maintain a significantly good end-to-end performance along with decent isolated recognition performance indicating that they can generalize well on both raw word-level inputs and processed words cropped from page-level inputs given that training is done only on the original IIT-Indic-HW-Words images. There is a performance drop in end-to-end recognition results compared to isolated results due to changed input images. The isolated evaluation takes the original word-level image as input whereas the end-to-end evaluation takes the processed word-level cropped image from the input page detected by our language agnostic \htd{} model.

\section{Conclusion}\label{sec:concl}

We address the gaps present in the evaluation of handwritten documents in Indic languages due to inconsistent comparison of multiple \htd{} and \htr{} models by introducing the \framework{} framework. \framework{}, being an end-to-end framework for page-level handwritten OCR helps to view this as a two-stage problem involving \htd{} followed by using a custom \htr{} model of the user's choice. Using our framework, we have demonstrated that we can evaluate each stage independently. We present the precision, recall, and F1 scores of \htd{} models as well as CRRs and WRRs of \htr{} models in an isolated and end-to-end manner using \framework{} framework, on ten Indic languages. Additionally, we also release \dataset, a meticulously curated, page-level Indic handwritten OCR dataset labeled for both detection and recognition. Thus, we address testing inconsistencies, data scarcity, and gaps encountered in page-level text recognition. Our future work aims to include script detection during word-level text detection. This allows us to dynamically select a language-specific \htr{} model without requiring the user to specify the script (or the language) as input. Furthermore, this will also facilitate the handwritten OCR of pages containing words in multiple scripts. Incorporating a robust language-agnostic \htr{} model could also be a promising direction further. We also hope to establish that the scores of these six HTR models and our analysis provide a reference point for future research.
\vspace{1cm}

\pagebreak

\renewcommand{\arraystretch}{1.6}
\begin{table*}[h]
\caption{Isolated performance of different \htr{} models on \dataset{} test set words}
\centering
\resizebox{\textwidth}{!}{%
\begin{tabular}{|c|cccccc|cccccc|}
\hline
\centering
{ \textbf{}} & \multicolumn{6}{c|}{{ \textbf{Character Recognition Rate (CRR \%)}}} & \multicolumn{6}{c|}{{ \textbf{Word Recognition Rate (WRR \%)}}} \\
\hline
\textbf{Language} & { \rotatebox{90}{CRNN}} & { \rotatebox{90}{MobileNet}} & { \rotatebox{90}{SAR}} & { \rotatebox{90}{VITSTR}} & { \rotatebox{90}{MASTER}} & { \rotatebox{90}{PARSEQ}} & { \rotatebox{90}{CRNN}} & { \rotatebox{90}{MobileNet}} & { \rotatebox{90}{SAR}} & { \rotatebox{90}{VITSTR}} & { \rotatebox{90}{MASTER}} & { \rotatebox{90}{PARSEQ}} \\
\hline
Bengali   & \textbf{94.92} & 84.95 & 60.39 & 84.71 & 84.86          & 92.83          & 68.53 & 28.68 & 03.19  & 35.41 & \textbf{72.50} & 64.48          \\
Gujarati  & \textbf{96.73} & 89.09 & 16.20 & 89.43 & 84.14          & 96.10          & 76.99 & 40.80 & 01.15  & 54.32 & 78.91          & \textbf{79.69} \\
Gurumukhi & \textbf{95.99} & 93.00 & 69.43 & 92.94 & 95.27          & 95.96          & 75.64 & 61.15 & 04.73  & 64.53 & \textbf{84.78} & 80.27          \\
Hindi     & 91.14          & 86.97 & 61.77 & 95.67 & 89.42          & \textbf{98.06} & 52.26 & 38.97 & 03.26  & 73.14 & 73.89          & \textbf{90.23} \\
Kannada   & \textbf{98.42} & 95.00 & 57.40 & 82.10 & 93.07          & 96.08          & 85.55 & 64.90 & 02.43  & 29.15 & \textbf{87.14} & 77.77          \\
Malayalam & \textbf{97.34} & 80.53 & 38.17 & 53.61 & 93.50          & 95.06          & 82.99 & 18.80 & 01.26  & 00.18  & \textbf{86.04} & 76.99          \\
Odia      & \textbf{95.58} & 88.49 & 43.80 & 67.16 & 74.02          & 94.15          & 71.32 & 41.01 & 03.06  & 04.64  & 50.75          & \textbf{72.32} \\
Tamil     & 98.11          & 94.45 & 43.56 & 91.51 & \textbf{98.33} & 97.21          & 83.69 & 58.27 & 00.91  & 51.76 & \textbf{89.09} & 81.38          \\
Telugu    & 95.84          & 85.69 & 57.57 & 51.63 & 91.14          & \textbf{97.21} & 63.35 & 23.45 & 01.02  & 00.01  & 74.15          & \textbf{80.62} \\
Urdu      & 92.31          & 86.30 & 62.60 & 33.75 & 86.83          & \textbf{95.00} & 70.40 & 53.05 & 12.88 & 00.06  & 79.22          & \textbf{86.60} \\
\hline
Overall  & 95.64 & 88.48 & 51.09 & 74.25 & 89.58 & \textbf{95.77} & 73.01 & 42.91 & 03.39 & 31.32 & 77.65 & \textbf{79.04}\\
\hline
\end{tabular}
}
\label{tab:isolated}
\end{table*}

\begin{table*}[h]
\caption{End-to-end performance of different \htr{} models on \dataset{} test set pages}
\centering
\resizebox{\textwidth}{!}{%
\begin{tabular}{|c|cccccc|cccccc|}
\hline
\centering
{ \textbf{}} & \multicolumn{6}{c|}{{ \textbf{Character Recognition Rate (CRR \%)}}} & \multicolumn{6}{c|}{{ \textbf{Word Recognition Rate (WRR \%)}}} \\
\hline
\textbf{Language} & { \rotatebox{90}{CRNN}} & { \rotatebox{90}{MobileNet}} & { \rotatebox{90}{SAR}} & { \rotatebox{90}{VITSTR}} & { \rotatebox{90}{MASTER}} & { \rotatebox{90}{PARSEQ}} & { \rotatebox{90}{CRNN}} & { \rotatebox{90}{MobileNet}} & { \rotatebox{90}{SAR}} & { \rotatebox{90}{VITSTR}} & { \rotatebox{90}{MASTER}} & { \rotatebox{90}{PARSEQ}} \\
\hline
Bengali   & \textbf{87.50} & 67.44 & 47.81 & 60.13 & 73.59          & 69.26 & \textbf{38.48} & 00.36 & 02.78 & 07.12 & 34.17 & 14.43 \\ 
Gujarati  & \textbf{83.96} & 63.60 & 11.88 & 52.68 & 71.48          & 65.47 & 30.85 & 00.03 & 00.35 & 02.32 & \textbf{32.12} & 11.72 \\ 
Gurumukhi & \textbf{88.40} & 74.20 & 51.49 & 63.79 & 87.28          & 74.90 & 43.60 & 00.81 & 06.22 & 18.67 & \textbf{52.74} & 22.76 \\ 
Hindi     & \textbf{77.42} & 64.58 & 46.13 & 57.02 & 76.65          & 57.84 & 17.35 & 00.53 & 01.58 & 03.82 & \textbf{34.03} & 04.20 \\ 
Kannada   & \textbf{94.62} & 74.90 & 48.04 & 48.76 & 89.07          & 70.97 & 59.01 & 00.46 & 00.08 & 12.72 & \textbf{62.24} & 13.14 \\ 
Malayalam & \textbf{94.43} & 66.89 & 40.87 & 43.36 & 90.09          & 76.12 & \textbf{62.27} & 00.35 & 00.01 & 03.41 & 61.56 & 21.97 \\ 
Odia      & \textbf{84.22} & 71.96 & 45.38 & 44.59 & 65.64          & 69.50 & \textbf{34.04} & 00.92 & 00.10 & 10.61 & 29.72 & 16.53 \\ 
Tamil     & \textbf{96.03} & 82.12 & 39.27 & 71.07 & 94.16          & 79.53 & \textbf{66.26} & 00.27 & 07.18 & 23.82 & 64.17 & 27.05 \\ 
Telugu    & \textbf{81.22} & 58.45 & 43.72 & 38.95 & 84.70          & 55.33 & 15.55 & 00.26 & 00.00 & 00.82 & \textbf{42.22} & 01.54 \\ 
Urdu      & 56.64          & 50.07 & 52.01 & 27.52 & \textbf{65.05} & 55.55 & 08.37 & 02.38 & 00.00 & 04.98 & \textbf{23.77} & 08.77 \\ 
\hline
Overall & \textbf{84.44} & 67.42 & 42.66 & 50.79 & 79.78 & 67.45 & 37.58 & 00.64 & 01.83 & 08.83 & \textbf{43.67} & 14.21 \\
\hline
\end{tabular}
}
\label{tab:end-to-end}
\end{table*}

\bibliographystyle{sn-basic}
\bibliography{references}

\end{document}